# Encoder-Decoder Network with Guided Transmission Map: Architecture

**Le-Anh Tran [1] and Dong-Chul Park [1]**
[1] Myongji University, Yongin, South Korea
E-mail: leanhtran@mju.ac.kr, parkd@mju.ac.kr

**Summary:** An insight into the architecture of the Encoder-Decoder Network with Guided Transmission Map (EDN-GTM), a novel and effective single image dehazing scheme, is presented in this paper. The EDN-GTM takes a conventional RGB hazy image in conjunction with the corresponding transmission map estimated by the dark channel prior (DCP) approach as inputs of the network. The EDN-GTM adopts an enhanced structure of U-Net developed for dehazing tasks and has shown state-of-the-art performances on benchmark dehazing datasets in terms of PSNR and SSIM metrics. In order to give an in-depth understanding of the well-designed architecture which largely contributes to the success of the EDN-GTM, extensive experiments and analysis from selecting the core structure of the scheme to investigating advanced network designs are presented in this paper.

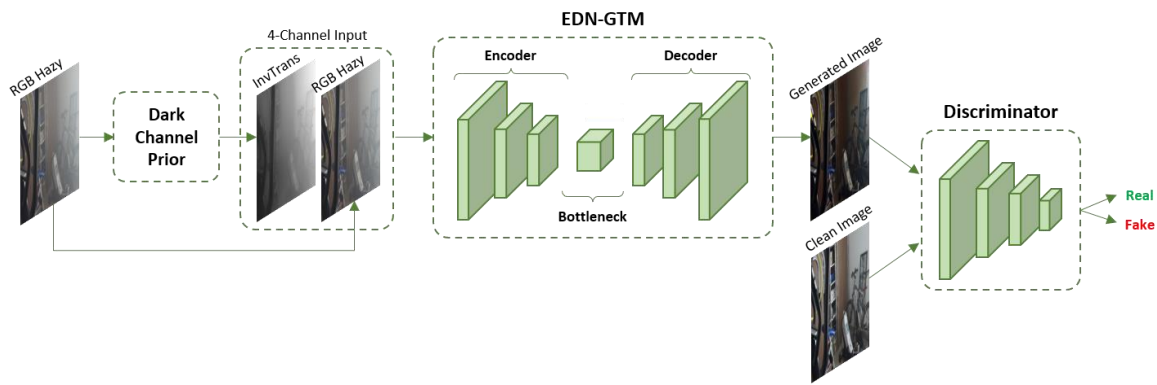

**Fig. 1.** The EDN-GTM scheme.

## 1. Introduction

Generally, haze can be considered to be one of the most fundamental phenomena causing image visibility degradation. Accurate estimation of the transmission map in a hazy image, however, has been a major obstacle in performing dehazing [1]. Numerous haze removal approaches have been proposed and most of them have achieved significant progress. Dehazing algorithms can be divided into two types: traditional methods and deep learning-based methods. Traditional approaches apply handcrafted models to perform haze removal tasks while deep learning-based schemes adopt convolutional neural networks (CNNs) in their systems. Both types have their own advantages.

In order to take the advantages of both types of dehazing algorithms, the Encoder-Decoder Network with Guided Transmission Map (EDN-GTM) has been proposed in our preliminary work which utilizes the transmission map extracted by the dark channel prior (DCP) as an additional input channel of a CNN model in order to achieve an improved dehazing performance [1]. To further provide an insight into the well-designed architecture that leads to the EDN-GTM's improvement, various experiments and analysis starting from selecting the core structure of the EDN-GTM scheme to evaluating the effect of every single modification on the network's performance are presented in this paper.

## 2. The EDN-GTM Scheme

The EDN-GTM scheme is illustrated in Fig. 1. In terms of the generator network, the EDN-GTM utilizes the transmission map estimated by DCP as an additional input channel of a U-Net-based generative network [2]. To further customize U-Net for dehazing task, the EDN-GTM applies three main modifications: 1) a spatial pyramid pooling (SPP) module is plugged into the bottleneck of U-Net; 2) ReLU activation is replaced with Swish activation; and 3) one convolution layer with the filter size of 3x3 is appended in each of the main convolution stages to increase the receptive field and capture more high-level features. In terms of the discriminator design, the encoding part of U-Net is utilized as the base network in order to encourage the discriminator to be equally capable of extracting and analyzing advanced features with the generator such that the two networks compete with each other to boost their performances.

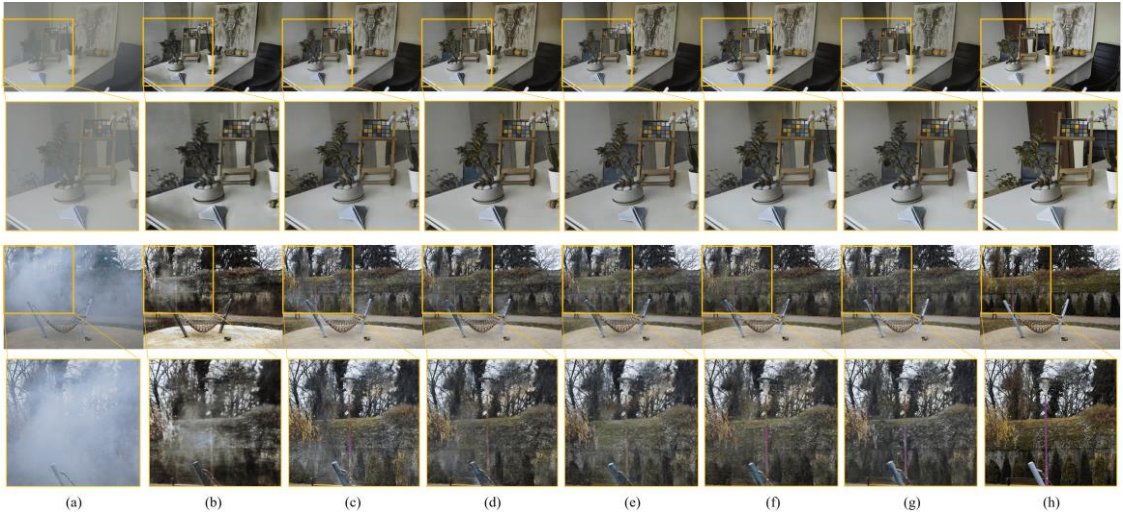

**Fig. 2.** Results across different network designs: (a) Input, (b) S-U-Net, (c) G-U-Net, (d) G-U-Net 4-C, (e) SPP G-U-Net 4-C (ReLU), (f) SPP G-U-Net 4-C (Swish), (g) EDN-GTM, and (h) Ground Truth (top: indoor, bottom: outdoor).

**Table 1.** Quantitative dehazing results of different network designs on 4 benchmark datasets.

| | S-U-Net | | G-U-Net | | G-U-Net 4-C | | SPP G-U-Net 4-C (ReLU) | | SPP G-U-Net 4-C (Swish) | | EDN-GTM | |
|---|---|---|---|---|---|---|---|---|---|---|---|---|
| **Dataset** | **PSNR** | **SSIM** | **PSNR** | **SSIM** | **PSNR** | **SSIM** | **PSNR** | **SSIM** | **PSNR** | **SSIM** | **PSNR** | **SSIM** |
| **I-HAZE** | 19.82 | 0.7805 | 21.04 | 0.8056 | 21.39 | 0.8115 | 22.20 | 0.8092 | 22.66 | **0.8311** | 22.90 | 0.8270 |
| **O-HAZE** | 20.96 | 0.7413 | 23.10 | 0.8099 | 23.15 | 0.8159 | 23.27 | 0.8174 | 23.43 | **0.8283** | 23.46 | 0.8198 |
| **Dense-HAZE** | 14.18 | 0.2954 | 14.96 | 0.4952 | 15.19 | 0.5062 | 15.43 | 0.5147 | **15.46** | **0.5359** | 15.43 | 0.5200 |
| **NH-HAZE** | 16.79 | 0.6368 | 19.18 | 0.6892 | 19.52 | 0.6877 | 19.73 | 0.7011 | 19.80 | 0.7064 | 20.24 | 0.7178 |

## 3. Architectural Analysis

In order to determine more acceptable architecture for the EDN-GTM, extensive experiments and analysis on different network configurations are conducted for evaluating the influence of each change of the network architecture on the dehazing performance. Based on the experimental results and evaluations, the optimal network design and the most effective data augmentation methods are selected to present the primary performance of the EDN-GTM scheme on benchmark dehazing datasets.

Overall, we have examined 12 different network configurations and 4 data augmentation methods: 1) core structures of the scheme including segmentation-like structure (S-U-Net) and generative structure (G-U-Net); 2) impact of the transmission map channel (G-U-Net 4-C); 3) advanced structures such as cross-stage partial module (CSP G-U-Net 4-C) and spatial pyramid pooling module (SPP G-U-Net 4-C); 4) attention mechanisms including spatial attention (SPP G-U-Net 4-C SAM) and channel attention (SPP G-U-Net 4-C CAM); 5) effects of various data augmentation methods such as random crop, horizontal flip, cutout, and mosaic; 6) effect of different activations including ReLU, Leaky ReLU, Swish, and Mish; and 7) impact of receptive field. Those configurations have been examined on 4 benchmark dehazing datasets (I-HAZE, O-HAZE, Dense-HAZE, and NH-HAZE [1]) in order to give comprehensive and concrete assessments of the EDN-GTM's design.

Typical visual results are illustrated in Fig. 2 while the quantitative evaluations are summarized in Table 1. Note that only 6 representative network designs, specifically S-U-Net, G-U-Net, G-U-Net 4-C, SPP G-U-Net 4-C (ReLU), SPP G-U-Net 4-C (Swish), and EDN-GTM, are chosen to provide the visual and quantitative results. The results demonstrate that a dramatic improvement in performance has been achieved when comparing the outcome of the original U-Net with that of the EDN-GTM scheme.

## 4. Results on Benchmark Datasets

In this section, we present the performances of the EDN-GTM scheme on various benchmark datasets for image dehazing tasks including I-HAZE, O-HAZE, Dense-HAZE, and NH-HAZE.

The quantitative results of the EDN-GTM scheme on I-HAZE and O-HAZE datasets compared with those of other modern dehazing approaches are first summarized in Table 2, while typical visual dehazing results produced by various dehazing methods on I-HAZE and O-HAZE datasets are presented in Fig. 3 (a) and Fig. 3 (b), respectively. As shown in Table 2, the EDN-GTM scheme achieves the best dehazing performance in PSNR metric (22.90 dB) while showing the second-best result in SSIM (0.8270) on I-HAZE dataset. On the other hand, on O-HAZE dataset, the EDN-GTM scheme produces the second-best result in PNSR measure (23.46 dB) while showing the best performance in terms of SSIM (0.8198). Further quantitative results of various haze removal algorithms on I-HAZE and O-HAZE datasets are summarized in Table 2, where the best and the second-best results are shown in red and blue colors, respectively.

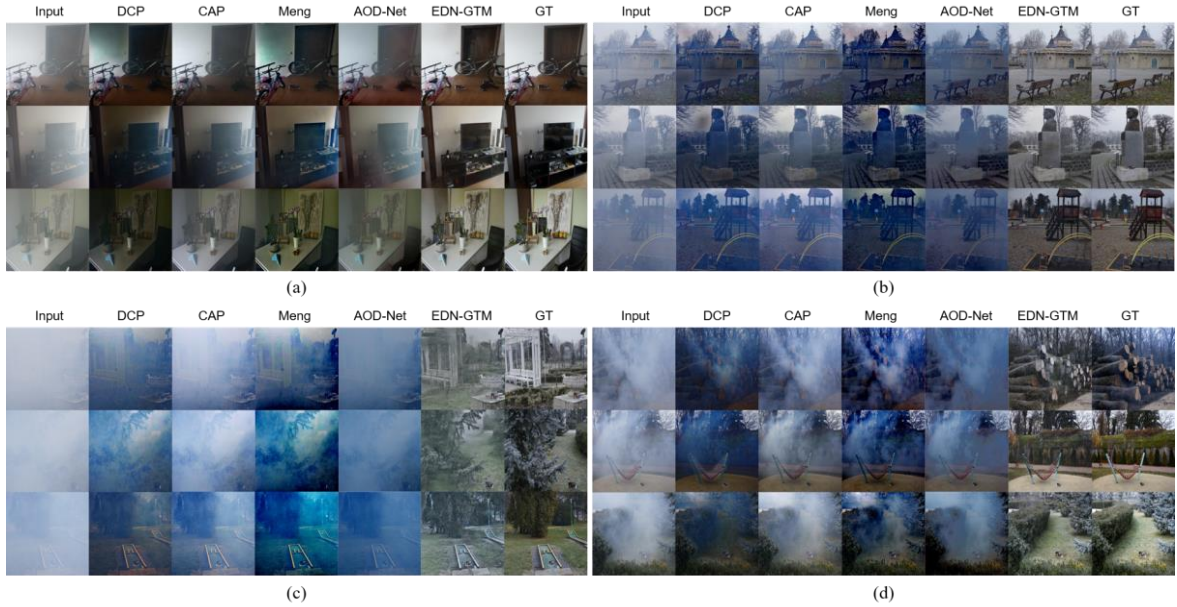

**Fig. 3.** Results of various single image dehazing methods (DCP [3], CAP [4], Meng *et al.* [5], AOD-Net [6], and EDN-GTM) on high-resolution datasets: (a) I-HAZE, (b) O-HAZE, (c) Dense-HAZE, and (d) NH-HAZE.

**Table 2.** Quantitative dehazing results on I-HAZE and O-HAZE datasets.

| Method | I-HAZE | | O-HAZE | |
|---|---|---|---|---|
| | PSNR | SSIM | PSNR | SSIM |
| DCP (TPAMI'10) [3] | 14.43 | 0.7516 | 16.78 | 0.6532 |
| CAP (TIP'15) [4] | 12.24 | 0.6065 | 16.08 | 0.5965 |
| MSCNN (ECCV'16) [7] | 15.22 | 0.7545 | 17.56 | 0.6495 |
| AOD-Net (ICCV'17) [6] | 13.98 | 0.7323 | 15.03 | 0.5385 |
| PPD-Net (CVPRW'18) [8] | 22.53 | 0.8705 | 24.24 | 0.7205 |
| EDN-GTM | 22.90 | 0.8270 | 23.46 | 0.8198 |

**Table 3.** Quantitative dehazing results on Dense-HAZE and NH-HAZE datasets.

| Method | Dense-HAZE | | NH-HAZE | |
|---|---|---|---|---|
| | PSNR | SSIM | PSNR | SSIM |
| DCP (TPAMI'10) [3] | 10.06 | 0.3856 | 10.57 | 0.5196 |
| DehazeNet (TIP'16) [9] | 13.84 | 0.4252 | 16.62 | 0.5238 |
| AOD-Net (ICCV'17) [6] | 13.14 | 0.4144 | 15.40 | 0.5693 |
| MSBDN (CVPR'20) [10] | 15.37 | 0.4858 | 19.23 | 0.7056 |
| AECR-Net (CVPR'21) [11] | 15.80 | 0.4660 | 19.88 | 0.7173 |
| EDN-GTM | 15.43 | 0.5200 | 20.24 | 0.7178 |

The quantitative results on Dense-HAZE and NH-HAZE datasets produced by the EDN-GTM scheme are also compared with those of other approaches and summarized in Table 3, where the best and the second-best results are indicated in red and blue colors, respectively. Note that Dense-HAZE and NH-HAZE datasets are more challenging than I-HAZE and O-HAZE datasets. On Dense-HAZE dataset, the EDN-GTM scheme provides the second-best result in PNSR measure (15.43 dB) while giving the best result in SSIM metric (0.5200). On NH-HAZE dataset, the EDN-GTM scheme convincingly achieves the best performances in both PSNR (20.24 dB) and SSIM (0.7178) measures. Some visual dehazing results from various methods on Dense-HAZE and NH-HAZE datasets are presented in Fig. 3 (c) and Fig. 3 (d), respectively.

As summarized in Table 2, Table 3, and Fig. 3, the EDN-GTM scheme is able to achieve favorable results on all the datasets examined in our experiments when compared with other conventional and recent dehazing methods. It implies that the EDN-GTM scheme's architecture has been well-designed as the scheme can perform efficiently on haze removal problems. Moreover, we notice that the transmission map plays a critical role in guiding the network to achieve promising results in image dehazing tasks.

## 5. Conclusions

An extensive analysis of the architectural features of the Encoder-Decoder Network with Guided Transmission Map (EDN-GTM), an effective single image dehazing scheme, is presented in this paper. In order to find an optimal architecture of the EDN-GTM, various features of architecture including core structure, transmission map channel, spatial pyramid pooling, activation functions, and receptive field variations are thoroughly studied. Various experiments examining the effect of every single change on the architectural features provide an in-depth understanding of the optimal design of the EDN-GTM scheme. Future work includes a study on the applicability of the EDN-GTM to an image pre-processing tool in hazy-weather circumstances to remove haze efficiently for autonomous driving systems [12,13,14].